\documentclass[11pt,letterpaper]{article}
\usepackage{emnlp2017}
\usepackage{times}
\usepackage{latexsym}
\usepackage[plain]{fancyref}
\usepackage{float}
\usepackage{algorithm}
\usepackage{algpseudocode}
\algnewcommand{\LineComment}[1]{\State \(\triangleright\) #1}
\newfloat{algorithm}{t}{lop}

\emnlpfinalcopy

\def\emnlppaperid{***}

\title{Neural machine translation for low-resource languages}

\author{Robert {\"O}stling \and J{\"o}rg Tiedemann \\
  {\tt robert@ling.su.se, jorg.tiedemann@helsinki.fi}}

\date{}

\begin{document}

\maketitle

\begin{abstract}
    Neural machine translation (NMT) approaches have improved the state of the
    art in many machine translation settings over the last couple of years,
    but they require large amounts of training data to produce sensible
    output.  We demonstrate that NMT can be used for low-resource languages as
    well, by introducing more local dependencies and using word alignments to
    learn sentence reordering during translation.  In addition to our novel
    model, we also present an empirical evaluation of low-resource
    phrase-based statistical machine translation (SMT) and NMT to investigate
    the lower limits of the respective technologies.  We find that while SMT
    remains the best option for low-resource settings, our method can produce
    acceptable translations with only 70~000 tokens of training data, a level
    where the baseline NMT system fails completely.
\end{abstract}

\section{Introduction}

Neural machine translation (NMT) has made rapid progress over the last few
years \citep{Sutskever2014sequenceto,Bahdanau2014nmt,Wu2016googlenmt},
emerging as a serious alternative to phrase-based statistical machine
translation \citep{Koehn2003pbsmt}.
Most of the previous literature perform empirical evaluations using training
data in the order of millions to tens of millions of parallel sentence pairs.
In contrast, we want to see how low you can push the training data requirement
for neural machine translation.\footnote{The code of our implementation is
    available at \url{http://www.example.com} (redacted for review, code
    submitted as nmt.tgz in the review system)}
To the best of our knowledge, there is no previous systematic treatment of
this question.
\citet{Zoph2016transferlearning} did explore low-resource NMT, but by assuming
the existence of large amounts of data from related languages.

\section{Low-resource model}

The current de-facto standard approach to NMT \citep{Bahdanau2014nmt} has two
components: a target-side RNN language model
\citep{Mikolov2010recurrent,Sundermeyer2012lstm}, and an encoder plus
attention mechanism that is used to condition on the source sentence.  While
an acceptable language model can be trained using relatively small amounts of
data, more data is required to train the attention mechanism and encoder.
This has the effect that a standard NMT system trained on too little data
essentially becomes an elaborate language model, capable of generating
sentences in the target language that have little to do with the source
sentences.

We approach this problem by reversing the mode of operation of the translation
model. Instead of setting loose a target-side language model with a weak
coupling to the source sentence, our model steps through the source sentence
token by token, generating one (possibly empty) chunk of the target sentence
at a time. The generated chunk is then inserted into the partial target
sentence into a position predicted by a reordering mechanism.
\Fref{tab:generation} demonstrates this procedure.  While reordering is a
difficult problem, word order errors are preferable to a nonsensical output.
We translate each token using a character-to-character model conditioned on
the local source context, which is a relatively simple problem since data is
not so sparse at the token level. This also results in open vocabularies for
the source and target languages.

\begin{table}
\begin{tabular}{llll}
    \textit{Input} & \multicolumn{2}{c}{\it Predictions} & \textit{State} \\
    Source & Target & Pos. & Partial hypothesis \\
    \hline \\
    I       & ich       & 1 & \textbf{ich} \\
    can     & kann      & 2 & ich \textbf{kann} \\
    not     & nicht     & 3 & ich kann \textbf{nicht} \\
    do      & tun       & 4 & ich kann nicht \textbf{tun} \\
    that    & das       & 3 & ich kann \textbf{das} nicht tun \\
\end{tabular}
\caption{Generation of a German translation of ``I can not do that'' in our
model.}
\label{tab:generation}
\end{table}

Our model consists of the following components, whose relation are also
summarized in Algorithm~\ref{alg:model}:
\begin{itemize}
    \item Source token encoder. A bidirectional
        Long Short-Term Memory (LSTM) encoder \citep{Hochreiter1997lstm} that
        takes a source token $w^s$ as a sequence of (embedded) characters and
        produces a fixed-dimensional vector $\textsc{src-tok-enc}(w^s)$.
    \item Source sentence encoder. A bidirectional LSTM encoder that
        takes a sequence of encoded source tokens and produces an encoded
        sequence $e^s_{1 \dots N}$.
        Thus we use the same two-level source sentence encoding scheme as
        \citet{Luong2016achievingopen} except that we do not use separate
        embeddings for common words.
    \item Target token encoder. A bidirectional LSTM encoder that takes
        a target token\footnote{Note that on the target side, a token may
            contain the empty string as well as multi-word expressions with
            spaces. We use the term \emph{token} to reflect that we seek to
            approximate a 1-to-1 correspondence between source and target
            tokens}
        $w^t$ and produces a fixed-dimensional vector
        $\textsc{trg-tok-enc}(w^t)$.
    \item Target state encoder. An LSTM which at step $i$ produces a target
        state
        $h_i = \textsc{trg-enc}(e^s_i \| \textsc{trg-tok-enc}(w^t_{i-1}))$
        given the $i$:th encoded source position and the $i-1$:th
        encoded target token.
    \item Target token decoder. A standard character-level LSTM language model
        conditioned on the current target state $h_i$, which produces a target
        token $w^t_i = \textsc{trg-tok-dec}(h_i)$ as a sequence of characters.
    \item Target position predictor. A two-layer feedforward network that
        outputs (after softmax) the probability of $w^t_i$ being inserted at
        position $k_i = j$ of the partial target sentence:
        $P(k_i=j) \propto \exp(\textsc{position}(
            h_{\textit{pos}(j)} \| h_{\textit{pos}(j+1)} \| h_i))$.
        Here, $\textit{pos}(j)$ is the position $i$ of the source token
        that generated the $j$:th token of the partial target hypothesis.
        This is akin to the attention model of traditional NMT, but less
        critical to the translation process because it does not directly
        influence the generated token.
\end{itemize}

\begin{algorithm}
    \caption{Our proposed translation model.}
    \label{alg:model}
    \begin{algorithmic}
        \Function{translate}{$w^s_{1 \dots N}$}
            \LineComment{Encode each source token}
            \ForAll{$i \in 1 \dots N$}
                \State $e^s_i \gets \textsc{src-tok-enc}(w^s_i)$
            \EndFor
            \LineComment{Encode source token sequence}
            \State $s_{1 \dots N} \gets \textsc{src-enc}(e^s_{1 \dots N})$
            \LineComment{Translate one source token at a time}
            \ForAll{$i \in 1 \dots N$}
                \LineComment{Encode previous target token}
                \State $e^t_i \gets \textsc{trg-tok-enc}(w^t_{i-1})$
                \LineComment{Generate target state vector}
                \State $h_i \gets \textsc{trg-enc}(e^t_i \| s_i)$
                \LineComment{Generate target token}
                \State $p(w^t_i) \sim \textsc{trg-tok-dec}(h_i)$
                \LineComment{Predict insertion position}
                \State $p(k_i) \sim \textsc{position}(h_{1 \dots i}, k_{1 \dots i-1})$
            \EndFor
            \LineComment{Search for a high-probability target}
            \LineComment{sequence and ordering, and return it}
        \EndFunction
    \end{algorithmic}
\end{algorithm}

\section{Training}
\label{sec:training}

Since our intended application is translation of low-resource languages, we
rely on word alignments to provide supervision for the reordering model.  We
use the \textsc{efmaral} aligner \citep{Ostling2016efmaral}, which uses a
Bayesian model with priors that generate good results even for rather small
corpora. From this, we get estimates of the alignment probabilities
$P_f(a_i=j)$ and $P_b(a_j=i)$ in the forward and backward directions,
respectively.

Our model requires a sequence of source tokens and a sequence of target tokens
of the same length. We extract this by first finding the most confident 1-to-1
word alignments, that is, the set of consistent pairs $(i,j)$ with maximum
$\prod_{(i,j)} P_f(a_i=j) \cdot P_b(a_j=i)$.
Then we use the source sentence as a fixed point, so that the final training
sequence is the same length of the source sentence. Unaligned
source tokens are assumed to generate the empty string, and source tokens
aligned to a target token followed by unaligned target tokens are assumed to
generate the whole sequence (with spaces between tokens).
While this prohibits a single source token to generate multiple
dislocated target words (e.g., standard negation in French), our
intention is that this constraint will result in overall better translations
when data is sparse.

Once the aligned token sequences have been extracted, we train our model using
backpropagation with stochastic gradient descent.
For this we use the Chainer library \citep{Tokui2015chainer},
with Adam \citep{Kingma2014adam} for optimization. We use early stopping based
on cross-entropy from held-out sentences in the training data. For the smaller
Watchtower data, we stopped after about 10 epochs, while about 25--40 were
required for the 20\% Bible data.
We use a dimensionality of 256 for all layers except the character embeddings
(which are of size 64).

\section{Baseline systems}

In addition to our proposed model, we two public translation systems: Moses
\citep{Koehn2007moses} for phrase-based statistical machine translation (SMT),
and HNMT\footnote{\url{https://github.com/robertostling/hnmt}} for neural
machine translation (NMT). For comparability, we use the same word alignment
method \citep{Ostling2016efmaral} with Moses as with our proposed model (see
\Fref{sec:training}).

For the SMT system, we use 5-gram modified Kneser-Ney language models
estimated with KenLM \citep{Heafield2013scalablemodified}. We symmetrize
the word alignments using the \textsc{grow-diag-final} heuristic.
Otherwise, standard settings are used for phrase extraction and estimation of
translation probabilities and lexical weights. Parameters are tuned using
MERT \citep{Och2003minimumerror} with 200-best lists.

The baseline NMT system, HNMT, uses standard attention-based translation
with the hybrid source encoder architecture of \citet{Luong2016achievingopen} 
and a character-based decoder. We run it with parameters comparable to those
of our proposed model: 256-dimensional word embeddings and encoder LSTM,
64-dimensional character embeddings, and an attention hidden layer size of 256.
We used a decoder LSTM size of 512 to account for the fact that this model
generates whole sentences, as opposed to our model which generates only small
chunks.

\section{Data}

The most widely translated publicly available parallel text is the
Bible, which has been used previously for multilingual NLP
\citep[e.g.][]{Yarowsky2001inducingtext,Agic2015bitof}. 
In addition to this, the Watchtower magazine is also publicly available and
translated into a large number of languages. Although generally containing
less text than the Bible, \citet{Agic2016multilingual} showed that its more
modern style and consistent translations can outweigh this disadvantage for
out-of-domain tasks.
The Bible and Watchtower texts are quite similar, so we also evaluate on data
from the WMT shared tasks from the news domain (\texttt{newstest2016} for
Czech and German, \texttt{newstest2008} for French and Spanish).
These are four languages that occur in all three data sets, and we use them
with English as the target language in all experiments.

The Watchtower texts are the shortest, after removing 1000 random sentences
each for development and test, we have 62--71 thousand tokens in each language
for training. For the Bible, we used every 5th sentence in order to get a
subset similar in size to the New Testament.\footnote{The reason we did not
    simply use the New Testament is because it consists largely of four
    redundant gospels, which makes it difficult to use for machine translation
    evaluation.}
After removing 1000 sentences for development and test, this yielded
130--175 thousand tokens per language for training.

\section{Results}

\begin{table*}[t]
    \begin{tabular}{lp{+.8\textwidth}}
        \hline\\
        Source & pues bien , la biblia da respuestas satisfactorias . \\
        Reference &
            the bible provides satisfying answers to these questions . \\
        SMT & well , the bible gives satisfactorias answers . \\
        HNMT & jehovah 's witness . \\
        Our & the bible to answers satiful . \\
        \hline\\
        Source & 4 , 5 . ¿ cuáles son algunas de las preguntas más importantes
        que podemos hacernos , y por qué debemos buscar las respuestas ? \\
        Reference & 4 , 5 . what are some of the most important questions we
        can ask in life , and why should we seek the answers ? \\
        SMT & 4 , 5 . what are some of the questions that matter most that we
        can make , and why should we find answers ? \\
        HNMT & 4 , 5 . what are some of the bible , and why ? \\
        Our & , 5 . what are some of the special more important that can , and
        we why should feel the answers \\
        \hline\\
    \end{tabular}
    \caption{Example translations from the different systems (Spanish-English;
        Watchtower test set, trained on Watchtower data).}
    \label{tab:examples}
\end{table*}

\Fref{tab:results} summarizes the results of our evaluation, and
\Fref{tab:examples} shows some example translations.
For evaluation we use the BLEU metric \citep{Papineni2002bleu}.
To summarize, it is clear that SMT remains the best method for low-resource
machine translation, but that current methods are not able to produce
acceptable general machine translation systems given the parallel data
available for low-resource languages.

Our model manages to reduce the gap between phrase-based and neural machine
translation, with BLEU scores of 9--17\% (in-domain) using only about 70~000
tokens of training data, a condition where the traditional NMT system is
unable to produce any sensible output at all.
It should be noted that due to time constraints, we performed inference with
greedy search for our model, whereas the NMT baseline used a beam search with
a beam size of 10.

\begin{table}
    \begin{tabular}{llrrr}
         & & \multicolumn{3}{c}{BLEU (\%)} \\
        Test & Source & SMT & HNMT & Our \\
        \hline
        \multicolumn{5}{c}{Trained on 20\% of Bible} \\
        \hline
        Bible & German & 25.7  & 7.9 & 10.2  \\
        Bible & Czech & 24.2  & 5.5 & 9.3  \\
        Bible & French & 39.7  & 19.8 & 25.7  \\
        Bible & Spanish & 22.5  & 3.9 & 9.3  \\
        Watchtower & German & 9.2 & 1.3 & 4.7 \\
        Watchtower & Czech & 7.5 & 0.7 & 3.5 \\
        Watchtower & French & 12.3 & 3.1 & 6.6 \\
        Watchtower & Spanish & 12.5 & 0.5 & 5.9 \\
        News & German & 4.1 & 0.1 & 1.7 \\
        News & Czech & 7.1 & 0.0 & 1.0 \\
        News & French & 9.0 & 0.0 & 2.4 \\
        News & Spanish & 6.5 & 0.0 & 1.7 \\
        \hline
        \multicolumn{5}{c}{Trained on Watchtower} \\
        \hline
        Bible & German & 7.7 & 0.3 & 3.7  \\
        Bible & Czech & 5.5 &  0.2& 1.8 \\
        Bible & French & 16.3 & 0.6  & 10.1 \\
        Bible & Spanish & 8.9 & 0.3 & 4.7 \\
        Watchtower & German & 27.6 & 2.5 & 11.2\\
        Watchtower & Czech & 26.3 & 1.6 & 8.7 \\
        Watchtower & French & 29.3 & 2.5 & 13.5\\
        Watchtower & Spanish & 35.7 & 3.0 & 17.0\\
        News & German & 9.5 & 0.1 & 1.8 \\
        News & Czech & 5.4 &  0.0& 0.8\\
        News & French & 9.2 & 0.0 & 2.5 \\
        News & Spanish & 9.4 & 0.0 & 1.8 \\
     \end{tabular}
     \caption{Results from our empirical evaluation.}
     \label{tab:results}
\end{table}

\section{Discussion}

We have demonstrated a possible road towards better neural machine translation
for low-resource languages, where we can assume no data beyond a small
parallel text. In our evaluation, we see that it outperforms a standard NMT
baseline, but is not currently better than the SMT system. In the future, we
hope to use the insights gained from this work to further explore the
possibility of constraining NMT models to perform better under severe data
sparsity. In particular, we would like to explore models that preserve more of
the fluency characteristic of NMT, while ensuring that adequacy does not
suffer too much when data is sparse.

%
%

\bibliography{emnlp17}
\bibliographystyle{emnlp_natbib}

\end{document}